\documentclass[12pt]{article}
\usepackage{frExamplee}
\usepackage{graphicx}
\usepackage{apalike}
\usepackage{setspace}

\usepackage{relsize}
\usepackage{xspace}
\usepackage{color}
\usepackage{graphics}
\usepackage{graphicx}

\usepackage{amsmath}
\DeclareMathOperator*{\argmax}{argmax}
\DeclareMathOperator*{\argmin}{argmin}
\usepackage{amsfonts}
\usepackage{amsmath}
\usepackage{amssymb}

\usepackage{subcaption}
\usepackage{float}
\usepackage{todonotes}
\usepackage{url} 
\usepackage{multirow}
\usepackage{hhline}

\graphicspath{{./figures/}} %

\newcommand{\acronym}[1]{{#1}\xspace}
\newcommand{\SLAM}{\acronym{\mbox{SLAM}}}
\newcommand{\CSLAM}{\acronym{\mbox{C-SLAM}}}

\newcommand{\norm}[1]{\left\lVert#1\right\rVert}

\title{Towards Collaborative Simultaneous Localization and Mapping: a Survey of the
Current Research Landscape}

\author{Pierre-Yves Lajoie \\
        Dept. Computer and Software Engineering \\
        Polytechnique Montréal \\
        \texttt{pierre-yves.lajoie@polymtl.ca} \\
        \And 
        Benjamin Ramtoula \\
        Oxford Robotics Institute, \\
        Dept. Engineering Science \\
        University of Oxford \\
        \texttt{benjamin@robots.ox.ac.uk} \\
        \And
        Fang Wu  \\
        Desay SV Automotive\\
        \texttt{fang.wu@desay-svautomotive.com} \\
        \And
        Giovanni Beltrame \\
        Dept. Computer and Software Engineering \\
        Polytechnique Montréal \\
        \texttt{giovanni.beltrame@polymtl.ca} \\
}

\begin{document}

\maketitle

\begin{abstract}
  Motivated by the tremendous progress we witnessed in recent years,
  this paper presents a survey of the scientific literature on the
  topic of Collaborative Simultaneous Localization and Mapping
  (\CSLAM), also known as multi-robot SLAM.  With fleets of
  self-driving cars on the horizon and the rise of multi-robot systems
  in industrial applications, we believe that Collaborative \SLAM will
  soon become a cornerstone of future robotic applications. In this
  survey, we introduce the basic concepts of \CSLAM and present a
  thorough literature review.  We also outline the major challenges
  and limitations of \CSLAM in terms of robustness, communication, and
  resource management.  We conclude by exploring the area's current
  trends and promising research avenues. 
\end{abstract}

\section{Introduction}

Collaborative Simultaneous Localization and Mapping (\CSLAM), also
known as multi-robot \SLAM, has been studied extensively with early
techniques dating back as far as the early
2000s (e.g.~\cite{jenningsCooperativeRobotLocalization1999,foxProbabilisticApproachCollaborative2000,thrunProbabilisticOnLineMapping2001,williamsMultivehicleSimultaneousLocalisation2002,fenwickCooperativeConcurrentMapping2002}).
These techniques were introduced only a short time after the inception of
single-robot \SLAM by researchers who were already envisioning collaborative
perception of the environment.  Although they were small-scale proofs
of concept, they laid the foundations that still shape the field
nowadays.

After years of confinement to laboratory settings, \CSLAM technologies
are finally coming to fruition into industry applications, ranging from
warehouse management to fleets of self-driving cars.  Those long
awaited success stories are a strong indicator that \CSLAM
technologies are poised to permeate other fields such as marine
exploration
\cite{paullDecentralizedCooperativeTrajectory2014,bonin-fontMultiRobotVisualGraphSLAM2020},
cooperative object transportation
\cite{riouxCooperativeSLAMbasedObject2015}, or search and rescue
operations
\cite{tianSearchRescueForest2020,leeEfficientRescueSystem2020}.

\SLAM is the current method of choice to enable autonomous navigation,
especially in unknown and GPS-denied environments.  \SLAM provides an
accurate representation of the robot surroundings which can in turn
enable autonomous control and decision making.  Similarly, in
multi-robot systems, \CSLAM enables collaborative behaviors by
building a collective representation of the environment and a shared
situational awareness.%
Moreover, many ambitious applications remain for multi-robot systems,
such as the exploration of other planets
\cite{vitugCooperativeAutonomousDistributed2021,ebadiDARESLAMDegeneracyAwareResilient2021}.  
To reach those moonshot goals, ongoing trends in
the research community aim to push the boundaries of multi-robot
systems towards increasingly larger teams, or swarms of robots
\cite{beniSwarmIntelligenceSwarm2004,kegeleirsSwarmSLAMChallenges2021},
which potentially allow parallel operations that are more efficient
and versatile.  
However, this is still largely uncharted territory
since current multi-robot applications either involve very few robots
or rely upon large amounts of centralized computation in server
clusters.  
Current \CSLAM techniques are no exception. They are prone to deteriorated
performance when the team size increases above a few robots, and could
be infeasible when minimal or no prior information is available about
the operating environment.  

Even though \CSLAM-enabled swarms of robots are still far from reality,
\CSLAM remains a useful tool when operating as few as two autonomous robots.
In exploration and mapping applications, even small teams can yield a
significant boost in performance compared to a single robot
system \cite{simmonsCoordinationMultiRobotExploration2000}. 
Notably, autonomous mapping using \CSLAM has recently received
a lot of attention due to the latest DARPA Subterranean Challenge
\cite{darpaDARPASubterraneanChallenge2020} and its potential
applications in space technologies
\cite{bezouskaDecentralizedCooperativeLocalization2019}.

Thus, this paper presents a survey of the relevant literature on the
topic of \CSLAM, aiming to give a complete overview of the main
concepts, current developments, open challenges, and new trends in the
field.  We hope it will help new as well as established researchers to
evaluate the state-of-the-art and offer valuable insights to guide
future design choices and research directions.  Compared to previous
reviews
\cite{saeediMultipleRobotSimultaneousLocalization2016,roneMappingLocalizationMotion2013},
this paper provides an update on the tremendous progress in the past
five years. 
In particular, we delve into the major advances towards the deployment of
complete \CSLAM systems outside closely monitored laboratory
environments, and we address the specific challenges of the different submodules
(i.e., front-end, back-end, etc.).
We also focus on the emergent trends and new opportunities coming from adjacent
fields of research (e.g. deep learning, edge computing, etc.).
This paper aims for a broader overview of the field than
surveys covering specific \CSLAM subproblems such as map merging
\cite{leeSurveyMapMerging2012}, practical implementations
\cite{kshirsagarSurveyImplementationMultiRobot2018}, particle filter
techniques \cite{guptaSurveyMultirobotParticle2019}, vision-based
techniques \cite{zouCollaborativeVisualSLAM2019}, and search and
rescue applications \cite{queraltaCollaborativeMultiRobotSearch2020}.

\subsection{Outline}
The rest of this paper consists of seven sections covering
the main \CSLAM subfields of research presented in
Table~\ref{tab:cslam_fields_of_research}: Section
\ref{sec:slam} presents an overview of the single robot \SLAM problem;
Section \ref{sec:cslam} explains the core differences with \CSLAM;
Section \ref{sec:frontend} explores the different modules of the \CSLAM
front-end and their challenges;
Section \ref{sec:inference} introduces the \CSLAM back-end and discusses the
different inference techniques; 
Section \ref{sec:system-challenges} looks into important system-level challenges.
Section \ref{sec:benchmarking} discusses the available benchmarking datasets;
Section \ref{sec:trends} presents open problems and ongoing trends in the fields;
and Section
\ref{sec:conclusion} concludes the survey and discusses future
research avenues.
\begin{table}
    \centering

    \begin{tabular}{|c|l|}
        \hline
        \multirow{3}{*}{\SLAM}               &   Odometry \\ 
                                             &   Intra-Robot Loop Closures \\ 
                                             &   Pose Estimation \\ 
                                             \hhline{==}
        \multirow{3}{*}{\CSLAM Front-End}    &   Direct Inter-Robot Loop Closures \\ 
                                             &   Indirect Inter-Robot Loop
                                             Closures \\ 
                                             &  Heterogeneous Sensing \\
                                             \hhline{==}
        \multirow{4}{*}{\CSLAM Back-End}     &   Extended Kalman Filters \\ 
                                             &   Particle Filters \\ 
                                             &   Pose Graph Optimization \\ 
                                             &   Perceptual Aliasing Mitigation \\  
                                             \hhline{==}
        \multirow{2}{*}{System-Level Challenges}    &   Map Representation \\  
                                             &   Communication Constraints \\ 
                                             \hhline{==}
        \multirow{5}{*}{Open Problems}  &   Resilient Inter-Robot Communication \\
                                             &   Managing Limited Computation Resources \\
                                             &   Adapting to Dynamic Environments \\ 
                                             &   Active \CSLAM \\ 
                                             &   Semantic \CSLAM \\ 
                                             &   Augmented Reality \\   
                                             \hline
    \end{tabular}

    \caption{Collaborative Simultaneous Localization and Mapping Subfields of Research}
    \label{tab:cslam_fields_of_research}
\end{table} %
\section{What is SLAM?}
\label{sec:slam}

At its core, \SLAM is a joint estimation of a robot's state and a
model of its surrounding environment, with the key assumption that a
moving robot performs the data collection sequentially. On one hand,
the robot's state comprises its pose (position and orientation) and
possibly other quantities such as sensors' calibration parameters.  On
the other hand, the environment model (i.e., the \textit{map}) consists of
representations of landmarks, built with processed data from the
robot's exteroceptive sensors such as cameras or lidars. This makes
\SLAM an essential part of many applications that require building an
accurate map of the surrounding environment, whether it be for
collision-free navigation, scene understanding, or visual inspection
by a remote human operator.  Since dead-reckoning approaches
(e.g. IMU, wheel or visual odometry) drift over time due to noise
accumulation, the environment map in \SLAM is also used internally to
correct the robot trajectory when known areas are re-visited.  The
recovered links between previously visited locations are called loop
closures.  \SLAM is useful when neither an a priori map nor
localization information are available, when a map needs to be
built, or long-term accurate localization estimates are required.
Common scenarios include robotics applications without external
positioning systems, such as the exploration of unknown indoor
environments, caves, mines, or other planets.

\subsection{Single-Robot SLAM problem}
Formally, the overall goal of \SLAM is to maximize the posterior of
the map and robot state.  We can formulate this with the state
variables $X$ of both the landmarks (map) and the robot, and the set
of measurements $Z$ acquired by the moving robot
\cite{thrunProbabilisticRoboticMIT2005}:
\begin{equation}
    p({X}|{Z})
    \label{eq:slam-posterior}
\end{equation}
This estimation problem is solved by either updating the current
state at each time step given the new observations (i.e., filtering) or optimizing
over the whole trajectory and past observations (i.e., smoothing).

Although filtering in \SLAM is still an active research topic, current
state-of-the-art techniques are mostly based on
smoothing \cite{cadenaPresentFutureSimultaneous2016,rosenAdvancesInferenceRepresentation2021a}. 
The common formulation for smoothing
techniques is a \textit{Maximum A Posteriori} (\textit{MAP})
estimation problem that leverages the moving robot assumption by
introducing a prior distribution (e.g. obtained by odometry) over the robot
trajectory.

Thus, the \SLAM problem for a single robot, designated with the lower case
letter $\alpha$, can be expressed
as finding $X_{\alpha}^{*}$, the solution of the \textit{MAP} problem:
\begin{equation}
    X_{\alpha}^{*} \doteq \argmax_{{X_{\alpha}}} p({X_{\alpha}}|{Z_{\alpha}}) = 
    \argmax_{{X_{\alpha}}} p({Z_{\alpha}}|{X_{\alpha}})p({X_{\alpha}})
    \label{eq:slam-problem}
\end{equation} 
The decomposition of the posterior distribution is obtained with Bayes'
theorem: $p({Z_{\alpha}}|{X_{\alpha}})$ is the likelihood of the
measurements $Z_{\alpha}$ given a certain $X_{\alpha}$, and
$p({X_{\alpha}})$ is the prior distribution of the robot motion
state. Intuitively, the \SLAM problem finds the set of state variables
(environment landmarks and robot poses) $X_{\alpha}^{*}$ that is most
likely to produce the measurements $Z_{\alpha}$ given a prior
estimation $p({X_{\alpha}})$.

It is important to also note that \SLAM is closely related to the
well-studied problem
of bundle adjustment in \textit{Structure from Motion} for which we
refer the reader to \cite{ozyesilSurveyStructureMotion2017}.

\subsection{SLAM Systems Architecture}
\begin{figure}
    \centering
    \includegraphics[width=0.6\paperwidth,trim=0mm 0mm 0mm 0mm,clip]{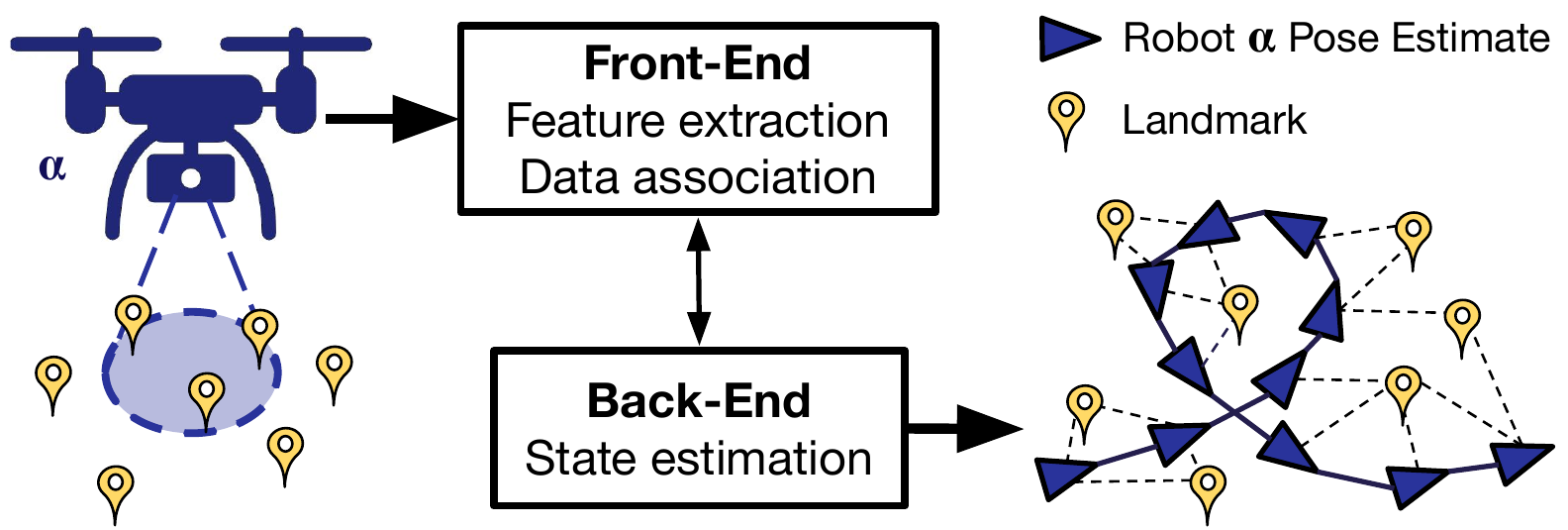}
    \caption{Single-robot \SLAM Overview}
    \label{fig:slam_overview}
\end{figure}

\SLAM systems are commonly divided into a front-end and a
back-end, each involving different fields of research. The
front-end is in charge of perception-related tasks, such as feature
extraction and data association which are both related to fields such
as computer vision and signal processing. The back-end 
produces the final state estimates using the front-end's outputs.
The back-end uses tools from the fields of optimization, probability
theory and graph theory.  In practice, the front-end processes the
sensor data to generate ego-motion, loop closure, and landmark measurements,
while the back-end performs the joint estimation of the map and the
robot state.  Figure~\ref{fig:slam_overview} provides an overview of a
common \SLAM structure in which the robot trajectory is represented as
a graph of poses at consecutive discrete times (i.e., a pose
  graph) and the map as a set of observed
landmarks \cite{cadenaPresentFutureSimultaneous2016}. In a 3D pose graph, the nodes are the robot poses
$[\mathbf{R}, \mathbf{t}] \in \mathrm{SE}(3)$ comprised of a rotation
matrix $\mathbf{R} \in \mathrm{SO}(3)$ and a translation
$\mathbf{t} \in \mathbb{R}^3$, and the edges represent the relative
measurements between the poses
\cite{barfootStateEstimationRobotics2017}.

Single-robot \SLAM still faces many challenges that consequently apply
to \CSLAM such as its long-term use, its robustness to perception
failures and incorrect estimates, or its need for performance
guarantees \cite{cadenaPresentFutureSimultaneous2016}.  To circumvent
those limitations in their specific settings, \SLAM and \CSLAM
developers often have to adapt the architecture and consider some
trade-offs between the sensors capabilities, the onboard computing
power, and available memory.

\section{What is Collaborative SLAM?}
\label{sec:cslam}

Many tasks can be performed faster and more efficiently by using
multiple robots instead of a single one.  Whether SLAM is used to
provide state estimation to support an application (e.g. estimate each
robot's position to plan for actions), or whether it is at the core of
the task (e.g. mapping an environment), it is beneficial and sometimes
necessary to extend SLAM solutions into coordinated \CSLAM algorithms
rather than performing single-robot SLAM on each robot.

\CSLAM algorithms aim to combine data collected on each individual
robot into globally consistent estimates of a common map and of each
robot's state. This coordination allows each robot to benefit from
experience of the full team, leading to more accurate localization and
mapping than multiple instances of single-robot \SLAM.  However, this
coordination introduces many new features and challenges inherent to
multi-robot systems.

\subsection{Multi-robot systems}
\label{sec:multi-rob}

In multi-robot systems, data collection and state estimation are no
longer entirely located on a single entity, so there is an inevitable
need for communication between the agents (i.e., robots, base stations,
etc.) which is the crux of the problem.

Moreover, multi-robot systems have additional properties to consider when
designing \CSLAM systems, and taxonomies can be defined to classify approaches
and highlight their benefits and tradeoffs. The taxonomy proposed in
\cite{dudekTaxonomy} presents considerations that are well suited to the
\CSLAM problem. It distinguishes approaches according to the following aspects:

\begin{description}
\item[Team size]
  The number of robots in the system. Larger teams usually
  perform tasks more efficiently but may be harder to coordinate.

\item[Communication range]
  Direct communication between robots is limited by
  their spatial distribution and the communication medium. In some cases, robots
  might be unable to communicate for long periods of time, while in others they
  might always be in range of another robot.

\item[Communication topology]
  The communication network topology affects how robots communicate with one
  another. For example, they might be limited to either broadcast or one-to-one
  messages.

\item[Communication bandwidth]
  The bandwidth of the communication channel affects what information robots can
  afford to share.

\item[System reconfigurability]
  The robots will move and are likely to change spatial configuration over time.
  This can affect the communication topology and bandwidth.

\item[Team unit processing ability]
  Individual robot's computational capability
  can affect the computation cost of \CSLAM approaches and the distribution of
  computation tasks.

\item[Team composition]
  Robots can be homogeneous or heterogeneous over several
  aspects such as locomotion methods and available sensors.
\end{description}
The main differences between most \CSLAM techniques in the literature lie in the
properties of the multi-robot system considered, especially their resource
management strategy. One subclass of multi-robot systems particularly relevant
to the future of \CSLAM are swarm robotics
systems~\cite{brambillaSwarmRobotics2013}, which are inspired by social animals.
Two main characteristics are required for swarm-compatibility in \CSLAM: robots'
sensing and communication capabilities must be local, and robots can not have
access to centralized control and/or to global knowledge.
Such systems would present considerable benefits: they would have robustness to
the loss of individual units, and they could scale well to large numbers of
robots.

\subsection{\CSLAM Problem definition}
When all robots' initial states are known or can be estimated, the \CSLAM
problem is a simple extension of the single-robot SLAM \textit{MAP} problem that
includes all the robots' states, measurements, and additional inter-robot
measurements linking different robots' maps. In a setup with two robots
($\alpha$, $\beta$), where $X_{\alpha}$ and $X_{\beta}$ are the state variables
from robot $\alpha$ and $\beta$ to be estimated, $Z_{\alpha}$ and $Z_{\beta}$
are the set of measurements gathered by robot $\alpha$ and $\beta$
independently, $Z_{\alpha\beta}$ is the set of inter-robot measurements linking
both robot maps containing relative pose estimates between one pose of robot
$\alpha$ and one of robot $\beta$ in their respective trajectories, and
$X_{\alpha}^{*}$, $X_{\beta}^{*}$ are the solutions, the problem can be
formulated as:

\begin{equation}
  \begin{aligned}
    (X_{\alpha}^{*}, X_{\beta}^{*}) &\doteq  \argmax_{{X_{\alpha},X_{\beta}}} p({{X_{\alpha},X_{\beta}}|Z_{\alpha},Z_{\beta},Z_{\alpha\beta}})\\
    &= \argmax_{{X_{\alpha},X_{\beta}}} p({Z_{\alpha},Z_{\beta},Z_{\alpha\beta}}|{X_{\alpha},X_{\beta}}) p({X_{\alpha},X_{\beta}})
  \end{aligned}
  \label{eq:cslam-problem-map}
\end{equation}

However, when the relative starting locations and orientations of the robots
cannot be determined, the initial guess of the robots states
$p({X_{\alpha},X_{\beta}})$ is not available. In that case, there are infinite
possible initial alignments between the multiple robot trajectories. Therefore,
in absence of a prior distribution, \CSLAM is reduced to the following
\textit{Maximum Likelihood Estimation} (MLE) problem.
\begin{equation}
    (X_{\alpha}^{*}, X_{\beta}^{*}) \doteq 
    \argmax_{{X_{\alpha},X_{\beta}}} p({Z_{\alpha},Z_{\beta},Z_{\alpha\beta}}|{X_{\alpha},X_{\beta}})
    \label{eq:cslam-problem}
\end{equation}

\begin{figure}
  \centering
  \includegraphics[width=\textwidth,trim=0mm 0mm 0mm 0mm,clip]{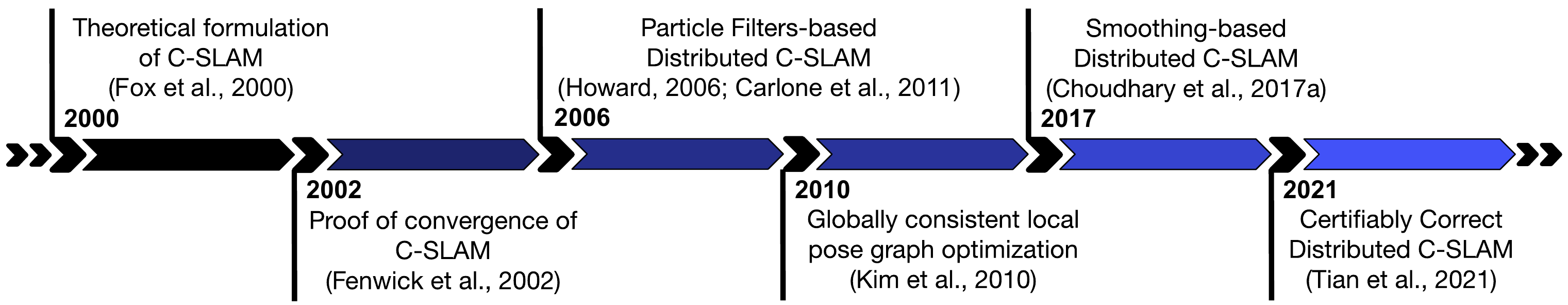}
  \caption{\CSLAM Problem Major Milestones}
  \label{fig:cslam_milestones}
\end{figure}

The \CSLAM problem formulation is still evolving to this day and
progress still needs to be made to achieve an efficient decentralized, distributed
and robust implementation.  To give some perspective, Figure
\ref{fig:cslam_milestones} presents some major milestones in
the evolution of the \CSLAM problem over time. More details on these milestone works are
provided in the following sections.

\subsection{Centralized, Decentralized and Distributed Systems}
An important distinction in \CSLAM, and in multi-robot systems in general,
is the difference
between the \textit{global} and \textit{local} perspectives.
The local perspective is the default point of view in single-robot
\SLAM.  Accordingly, the pose and map estimates are expressed
in an internal reference frame which is usually 
the starting location of the robot's mission.
 However, in
\CSLAM, one has to consider the global perspective of the system since
the pose and map of each robot need to be expressed in a
shared global reference frame.  This means that every landmark can be
expressed within the same coordinates system by every robot in the team.
Otherwise, shared information (e.g. position of observed landmarks) would
have no significance to the receiving robot due to the representation
being in another unknown local reference frame.  Establishing this
global reference frame using \CSLAM allows the robots to collectively
perceive the environment and benefit from each other's observations.

To achieve this global understanding, 
one could either solve \CSLAM
in a centralized or decentralized manner. 
In a centralized solution, the estimator has a global view
 of the entire team of robots: it performs the
estimation given perfect knowledge of the measurements of each robot. 
These measurements can be raw or preprocessed, and shared on demand 
depending on the communication limits.

\begin{figure}[H]
    \centering
    \begin{subfigure}[b]{0.49\columnwidth}
        \includegraphics[width=\columnwidth,trim=0mm 0mm 0mm 0mm,clip]{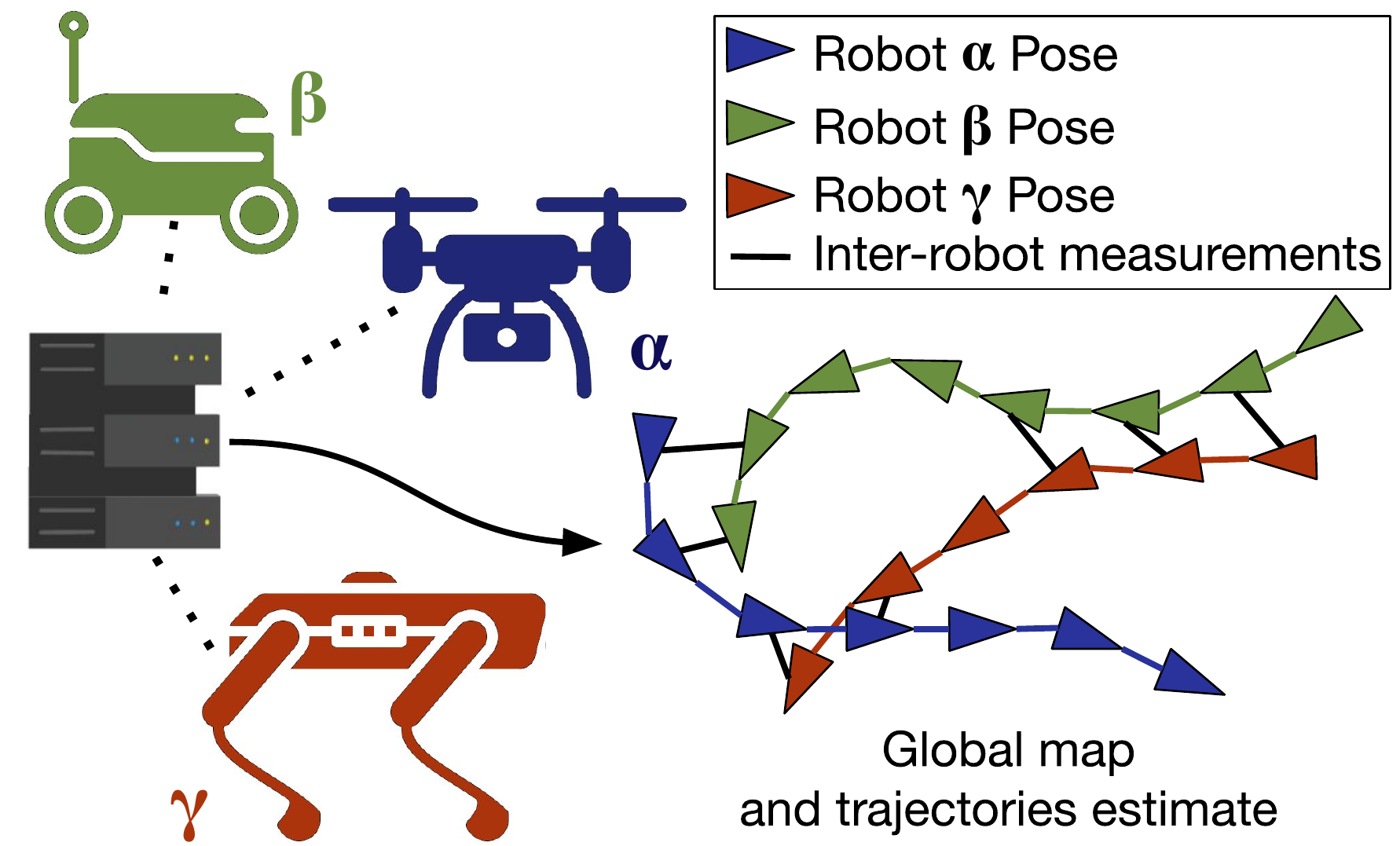}
        \caption{Centralized \CSLAM}
        \label{fig:global_cslam_overview}
    \end{subfigure}
    \begin{subfigure}[b]{0.49\columnwidth}
        \includegraphics[width=\columnwidth,trim=0mm 0mm 0mm 0mm,clip]{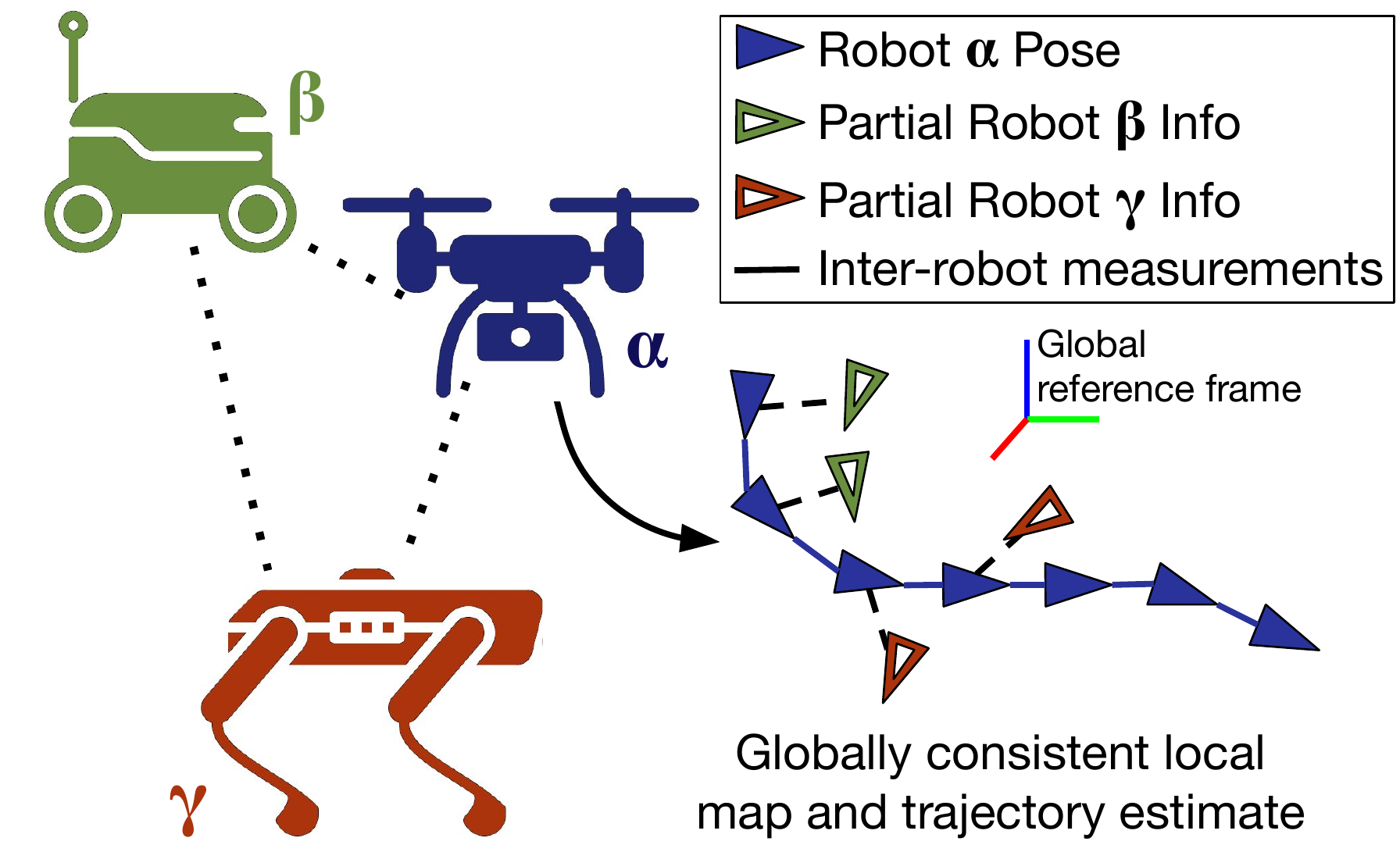}
        \caption{Decentralized \CSLAM}
        \label{fig:local_cslam_overview}
    \end{subfigure}
    \caption{Illustration of centralized and decentralized approaches to solve
    the \CSLAM estimation problem. The decentralized illustration is from the
    local perspective of robot $\alpha$.}
    \label{fig:global_local_cslam_overview}
\end{figure}

Unfortunately, due to communication constraints, solving centralized \CSLAM quickly
becomes intractable as the number of robots
increases \cite{saeediMultipleRobotSimultaneousLocalization2016}. 
Thus, a better solution for scalability is
to solve \CSLAM in a decentralized manner \cite{cieslewskiDataEfficientDecentralizedVisual2018}.
This means that each robot only has access to a local view comprised of
its own data and partial information from its neighbors.  
Therefore, decentralized systems
cannot solve the \CSLAM problem for all the robots at once, but
aim instead for local solutions on each robot that are consistent with their
neighbors.  
Then, iteratively and over time, with the robots gradually
improving their estimates given their neighbors' latest data, decentralized
techniques converge to local solutions that are mutually consistent
across the team of robots. 
So, upon convergence, the individual robots
reach a common understanding and their local maps are aligned with the
common (global) reference frame. 
Figure \ref{fig:global_local_cslam_overview} provides examples of the
\CSLAM problem and output in both perspectives.

Aside from the centralized/decentralized classification, a
system is distributed if the computation load is divided among the
robots.  The two notions are independent.  Therefore, a system could
be centralized and distributed at the same time, if, for example, each
robot performs parts of the computation, but a central node is
required to merge the individual results from all the robots \cite{leungCooperativeLocalizationMapping2012}.

\paragraph{Centralized C-SLAM}
Many seminal \CSLAM works are centralized and solve the
estimation problem in eq.~\ref{eq:cslam-problem} from the global
perspective. 
In those approaches, the robots are essentially reduced to mobile sensors 
whose data is collected and processed on a single computer.
Examples of centralized \CSLAM techniques
include
\cite{anderssonCSAMMultiRobotSLAM2008,kimMultipleRelativePose2010}
that gather all the robots' measurements at a central station to
compute the global map.  \cite{lazaroMultirobotSLAMUsing2013} improves
this solution by marginalizing unnecessary nodes in the local pose
graphs so only a few condensed measurements need to be shared to the
central computer.  Other centralized approaches
\cite{forsterCollaborativeMonocularSLAM2013,schmuckMultiUAVCollaborativeMonocular2017,schmuckCCMSLAMRobustEfficient2019}
perform \CSLAM with monocular cameras, successfully solving the
associated 3D estimation challenges, while 
\cite{loiannoCooperativeLocalizationMapping2015} focuses on
micro-aerial vehicles constraints.
\cite{deutschFrameworkMultirobotPose2016} proposes a framework to
reuse existing single robot \SLAM solutions for \CSLAM.  The same idea
is explored in \cite{liCORBSLAMCollaborativeVisual2018}, in which a
popular single-robot \SLAM technique \cite{murORB2} is converted into \CSLAM.
\cite{karrerCVISLAMCollaborativeVisualInertial2018,karrerGloballyConsistentVisualInertial2018}
integrate inertial measurements from IMUs in their centralized \CSLAM
systems.  
\cite{jimenezDecentralizedOnlineSimultaneous2018} proposes that the 
central node spreads the resulting
map across the robots to limit the memory usage.

Improving upon the pure centralized methods, some techniques do not rely on
a single computer, but can use different robots or base stations for
the computation.  This way, the system can adapt itself to the failure
of one node or communication link and complete the mission.  
A typical solution is to use replicated central servers
among the robots
\cite{baileyDecentralisedCooperativeLocalisation2011}.  

\paragraph{Decentralized C-SLAM}

Solving the \CSLAM problem in a decentralized manner is radically
different, but offers major benefits in terms of computation,
communication and privacy
\cite{choudharyDistributedMappingPrivacy2017,cieslewskiDataEfficientDecentralizedVisual2018}.
Such systems are usually
distributed and solve the estimation problem from
eq.~\ref{eq:cslam-problem} partially on each robot. 
As shown in Figure
\ref{fig:local_cslam_overview}, each robot computes its own local map
and uses partial information from other robots as well as inter-robot
measurements to achieve a local solution. 
Over several
iterations with its neighbors, 
each robot's resulting local solution converges to a
solution consistent with the global reference frame.  These techniques
mitigate communication and computation bottlenecks since the loads are spread
across the robot team \cite{pfingsthornScalableHybridMultirobot2008}.
Alternatively,
the full mapping data can be sent to every robot for redundancy and 
 a subset of robots can be designated for computation
\cite{saeediMultipleRobotSimultaneous2011,bressonConsistentMultirobotDecentralized2013,saeediOccupancyGridMap2015}.

As one would expect, decentralized and distributed techniques come with many additional
challenges that need to be tackled such as complex bookkeeping,
information double counting or synchronization issues.   %

\subsection{Complete \CSLAM Systems}

In \CSLAM, as well as in single-robot SLAM, the front-end handles
perception-related tasks and the back-end generates state estimates
using all measurements gathered.  However, in \CSLAM, the front-end
and back-end computations do not necessarily occur fully on a single
robot anymore depending on the sensing, communication, and estimation
strategies.
For example, in a centralized system, all robots could send their
sensor data directly to a single unit which would then perform the
front-end and back-end steps for the whole team.  While in a
decentralized and distributed setup, a robot could perform feature
extraction on its own and communicate with other robots for data
association and state estimation.  Every part of a \CSLAM system can
be subject to distribution or decentralization.
In addition, the front-end needs to
find links and relative measurements between the individual maps.
Therefore, the front-end must also perform data association to detect
and compute inter-robot loop closures, which will be detailed in
Section \ref{sec:frontend}.  It follows that the back-end must generate
estimates combining the individual and shared measurements as
described in Section \ref{sec:inference}.

In the recent years, several complete \CSLAM systems have been developed and
deployed in realistic scenarios. 
For example, some solutions deployed in large-scale environments during the DARPA Subterranean Challenge~\cite{hudson2021heterogeneous,aghaNeBulaQuestRobotic2021} led to the developments of new \CSLAM systems, such as the robust lidar-based approach of \cite{ebadiLAMPLargeScaleAutonomous2020}.
Alternatively, \cite{schmuckCOVINSVisualInertialSLAM2021} proposes a vision-based centralized \CSLAM system
incorporating inertial measurements, which has been tested with up to 12 robots in simulation.
In another line of work,
\cite{lajoieDOORSLAMDistributedOnline2020} presents a distributed and
decentralized system robust to spurious measurements, along with online
experiments on real robots, and a publicly available implementation.
A subsequent approach, detailed in \cite{tianKimeraMultiRobustDistributed2021},
puts together a completed decentralized and distributed \CSLAM system including
a novel robust distributed pose graph optimization back-end, and a front-end
producing globally consistent metric-semantic 3D mesh models of the explored environment. 
Those works are some of the best starting points for researchers
and engineers looking to implement, improve and deploy practical \CSLAM systems in
challenging conditions.

\section{\CSLAM Front-End}

\label{sec:frontend}

Although the division between the front-end and the back-end is
sometimes blurry due to the presence of feedback loops between the two
processes, a typical \CSLAM front-end is in charge of producing landmark
estimates, 
odometry measurements, and both intra-robot and inter-robot loop
closures.

Odometry measurements aim to capture the translation and rotation of a
robot from one time step to the next.  Common techniques measure wheel
movements, integrate from an IMU, and/or perform geometric matching
between consecutive images or laser-scans.  Intra-robot loop closures
are the measurements used by a \SLAM system to relocalize itself and
reduce its estimate error caused by odometry drift.  Using place
recognition, the system can detect previously visited locations and
compute relative measurements between them.  In other words,
intra-robot loop closures are estimates relating non-consecutive poses
in the robot trajectory that observed the same places.  Since the
computing of odometry and intra-robot loop closure measurements can be
fully done locally on each robot, the approaches used are the same in
both \SLAM and \CSLAM. Thus, we refer the reader to
\cite{mohamedSurveyOdometryAutonomous2019,cadenaPresentFutureSimultaneous2016,lowryVisualPlaceRecognition2016}
for surveys of the current techniques.

Conversely, inter-robot loop closures relate poses of different robots
trajectories.  They are the seams that stitch together the estimates
from multiple robots: they draw connections between the individual
robots' local maps to build the global understanding of the
environment.  Generating inter-robot loop closures is the main focus
of contributions to the front-end of \CSLAM systems, and key to ensure
consistency of the estimates.

\subsection{Direct vs Indirect Inter-Robot Loop Closures Measurements}
\label{subsec:loopclosures}

Inter-robot loop closures can be classified as direct or indirect
\cite{kimMultipleRelativePose2010}. Direct inter-robot loop closures
occur when two robots meet, and they are able to estimate their
current relative location with respect to each other through direct
sensing. Indirect inter-robot loop closures are produced by looking
back into maps to find partial overlaps for places that have been
visited by both robots.  Given these measurements, the robots
can estimate the relative transformation between their maps.  In
general, indirect inter-robot loop closures detection produce more
measurements and usually achieve a higher accuracy, but require more
communication and processing.  Indeed, the detection process is often
the communication bottleneck of \CSLAM given the large amount of data
required to compare landmarks between the individual local maps \cite{tardioliVisualDataAssociation2015}.

\subsubsection{Direct Inter-Robot Loop Closures}

The idea of direct inter-robot loop closures is to compute the
relative pose between two or more robots when they physically meet in
the same location. This is usually done through direct sensing of one
another.  For example, \cite{kimMultipleRelativePose2010} operated a
quadcopter and a ground robot and the latter was equipped with a
checkerboard pattern that could be detected by the quadcopter's
camera. \cite{zhouMultirobotSLAMUnknown2006} used a combination of
direct and indirect detection approaches, where colored cylinders were
installed to be detected by omnidirectional cameras.  In addition,
\cite{gentnerCooperativeSimultaneousLocalization2018,borosonInterRobotRangeMeasurements2020,caoVIRSLAMVisualInertial2021}
propose to replace visual loop closures by Ultra-Wide Band (UWB)
measurements from beacons onboard the robots. Given a few distance
measurements provided by the UWB sensors, the robots can estimate
their current relative pose with respect to each other and establish a
common reference frame.

\subsubsection{Indirect Inter-Robot Loop Closures}
\label{subsec:ind_loop_closures}
Indirect inter-robot loop closure detection is the extension of
single-robot loop closure detection to multiple maps.  In fact,
approaches to find indirect inter-robot loop closures often rely on
the same core algorithms as intra-robot loop closures.
The first challenge is the loop closures detection, which consists of
detecting overlaps between the individual maps.  
This task is usually
handled by a place recognition module which can efficiently compare
new observations against previous sections of the robots'
maps.
Following
place recognition matches, geometric estimation is performed to
compute the relative pose between the two places.

In the case of visual sensors, the place recognition problem has been
studied extensively \cite{lowryVisualPlaceRecognition2016}.  The
seminal work of visual bags of binary words
\cite{galvez-lopezBagsBinaryWords2012} is still popular, but newer
approaches based on deep learning, such as NetVLAD
\cite{arandjelovicNetVLADCNNArchitecture2018}, are more accurate and
data-efficient. Loop closure relative pose measurements can be
computed using visual features matching and multi-view geometry
\cite{hartleyMultipleViewGeometry2003}.

Finding inter-robot overlaps is a harder task with 3D point
clouds given the dense data that need to be shared and the lack of
expressive features to perform place recognition.  To that end,
compact and robust global point cloud descriptors
\cite{uyPointnetvlad} can be relied upon to compare point
clouds for place recognition. 
 Other approaches extract features from
the point cloud that can serve for place recognition while providing
initial guesses for later geometric alignments
\cite{ebadiDARESLAMDegeneracyAwareResilient2021}, or even directly
compute loop closure measurements \cite{dubeSegmatch}.  While
the classical \textit{Iterative Closest Point} method
\cite{beslMethodRegistration3D1992} is still commonly used in single
robot \SLAM to compute relative pose measurements between two matching
point clouds, it is not well suited for multi-robot operation due to
its reliance on a good initial guess, which is usually not available
between the robots' local maps.  
To handle the initialization problem, early work from \cite{olsonRealtimeCorrelativeScan2009}
presents a correlation-based algorithm that can be efficiently solved on a
GPU for real-time scan matching.
Another common solution is to
 use submaps matching for both stereo cameras
\cite{schusterMultirobot6DGraph2015,schulzSimultaneousCollaborativeMapping2019,duboisDenseDecentralizedMultiRobot2020}
and lidars
\cite{dubeOnlineMultirobotSLAM2017,ebadiDARESLAMDegeneracyAwareResilient2021}.
During this process, multiple laser scans or 3D point clouds are
clustered into submaps which can in turn be registered more efficiently.

\subsection{Heterogeneous Sensing}
\label{subsubsection:Heter_sensing}
In many applications, the teams of robots are composed of different
platforms equipped with different onboard sensors.  Heterogeneous
sensing comes with the additional challenge of matching map data in
different representation to perform relocalization and/or map fusion.
To this end, a recent study evaluated the repeatability of existing
keypoint detectors between data from stereo cameras and lidars For
example, when matching data from both stereo cameras and lidars, one
needs to chose repeatable 3D feature representations that are
consistent despite the differences in density and field-of-view
\cite{boroson3DKeypointRepeatability2019}.  Another approach is to use
an intermediate map representation that can be produced by different
kinds of sensors \cite{kochManagingEnvironmentModels2016}.  For
example, \cite{kaslinCollaborativeLocalizationAerial2016} proposes to
compare elevation maps that are invariant to sensor choice: lidars or
cameras.

\subsection{Non-Conventional Sensing}

While most \CSLAM techniques use the typical \SLAM sensors
(i.e., lidars and monocular, RGB-D, or stereo cameras), many recent research
works have explored the use of non-conventional sensors:
\cite{choiMultirobotMappingUsing2014} uses omnidirectional
(i.e., fish-eye) cameras,  \cite{waniekCooperativeSLAMSmall2015}
performs \CSLAM with event-based vision sensors, and
\cite{moralesInformationFusionStrategies2018} integrates ambient radio
signals (i.e., signals of opportunity) into their system.  In a
similar vein, \cite{liuCollaborativeSLAMBased2020} leverages existing
WiFi access points in most indoor environments to perform loop
closures based on their radio signal fingerprint.  
Alternatively, some approaches use only a few higher-level landmarks, such as
objects,
for tracking and place recognition~\cite{cunninghamDDFSAMFullyDistributed2010,choudharyMultiRobotObjectBased2017}.
This type of approach have regained popularity 
with the increasing performance of deep learning-based methods
in semantic segmentation as discussed in
Section~\ref{sec:semantics}.

\section{\CSLAM Back-End}
\label{sec:inference}

As mentioned before, the role of the \CSLAM back-end is to estimate
the state of the robot and the map given the front-end measurements.
The difference with single-robot \SLAM is the presence of inter-robot
measurements, the need to reach consensus, and the 
potential lack of an initial
guess since the global reference frame and the starting positions of
the robots are usually initially unknown.  Nevertheless, similar to
single-robot solvers, \CSLAM back-ends are roughly divided in two main
categories of inference techniques: filtering-based and
smoothing-based.  Although filtering-based approaches were the most
common among the early techniques (e.g. EKF
\cite{rekleitisProbabilisticCooperativeLocalization2003} and particle
filters \cite{madhavanDistributedCooperativeOutdoor2004}),
smoothing-based approaches quickly gained in popularity and are
currently considered as superior in most applications
\cite{strasdatVisualSLAMWhy2012}.  This section provides an overview
of the different categories of estimation workhorses for \CSLAM and
presents examples from the literature.

\subsection{Filtering-Based Estimation}

Filtering approaches are often characterized as online in the sense
that only the current robot pose is estimated and all previous poses
are marginalized out \cite{thrunProbabilisticRoboticMIT2005} at each
time step.  Consequently, the estimation of the posterior in
eq.~\ref{eq:slam-posterior} at time $t$ only depends on the posterior
at time $t-1$ and the new measurements.

The classical filtering technique for nonlinear problems (i.e., all
problems in robotics except trivial ones) is the Extended Kalman
Filter (EKF). It has been applied to \CSLAM in various ways among
which the information filter method presented in
\cite{thrunMultirobotSLAMSparse2005}.  In a nutshell, EKF are Gaussian
filters that circumvent the linear assumptions of Kalman filters
through linearization (i.e., local linear approximation); however, the
linearization process potentially leads to inconsistencies when the
noise is too large.  A major advantage of EKF techniques
\cite{thrunMultirobotSLAMSparse2005,sasaokaMultirobotSLAMInformation2016a,luftRecursiveDecentralizedCollaborative2016,schusterDistributedStereoVisionbased2019}
over smoothing techniques is that the covariance matrix is available
without requiring additional computation, which can be useful for
feature tracking or active exploration. For example, one could
prioritize the exploration in the most uncertain directions.  Yet, an
explicit covariance matrix is rarely required, so alternative
filtering techniques seek to avoid its computation, such as the smooth
variable structure filters approach presented in
\cite{demimCooperativeSLAMMultiple2017}.

Building on the EKF, Rao-Blackwellized Particle Filters (RBPF)
\cite{doucetRaoBlackwellisedParticleFiltering2000} are another
popular filtering approach for the \CSLAM problem. Techniques, such as
\cite{howardMultirobotSimultaneousLocalization2006}, use samples
(particles) to represent the posterior distribution
in eq.~\ref{eq:slam-posterior} and perform variable marginalization using
an EKF which drastically reduce the size of the sampling space.
\cite{carloneSimultaneousLocalizationMapping2011} extends on
\cite{howardMultirobotSimultaneousLocalization2006} and improves its
consistency while making it fully distributed.
\cite{gilMultirobotVisualSLAM2010} adapts RBPF to visual \CSLAM and
\cite{dorrCooperativeLongtermSLAM2016} showcases the potential of RBPF
\CSLAM for industrial applications.

It is important to note that, according to theoretical analysis results~\cite{mourikisAnalysisPositioningUncertainty2004}, reducing the number of relative position
measurements between the robots to a minimum, to limit communication and computation, 
only inflicts a small penalty on the localization performance.
It was also shown that the presence of even only one robot equipped with an
absolute positioning sensor is enough to bound the positioning uncertainty of
the whole team.
Additionally, analytical upper bounds can be computed
to predict the positioning uncertainty as function of the size of the explored
area, the number of landmarks, the number of robots, and
the accuracy of the onboard
sensors~\cite{mourikisPredictingPerformanceCooperative2006}.
Those theoretical results can be of great use in the design of a \CSLAM system. 

\subsection{Smoothing-Based Estimation}

Besides the linearization error, another drawback of filtering
techniques is that the marginalization of past pose variables leads to
many new links among the remaining variables. Indeed, the elimination
of each pose variable leads to interdependence between every landmark
variables to which it was connected. As a result, the variables become
increasingly coupled and this leads to more computation. 
This problem also affects smoothing approaches, but a clever ordering during
variable elimination can significantly reduce its impact on performance \cite{dellaertSquareRootSAM2006}.
Moreover, in
smoothing, there is less marginalization required which means that the
variables will stay sparsely connected. This sparsity is exploited by
modern solvers to yield significant speed-ups
\cite{strasdatVisualSLAMWhy2012}.  
In addition, unlike filtering-based
approaches, smoothing techniques improve their accuracy by revisiting
past measurements instead of only working from the latest estimate.
Hence, filtering techniques fell out of favor due to the better
performance of smoothing both in terms of accuracy and efficiency.  
Moreover, in the context of \CSLAM,
the sparsity reduces the amount of data to be exchanged during the
estimation process \cite{paullCommunicationconstrainedMultiAUVCooperative2015}.

In order to formalize the estimation problem solved by \CSLAM back-ends, 
we present a general smoothing formulation for
pose-graph \CSLAM with two robots $(\alpha, \beta)$ in which the map
landmarks are marginalized into odometry and loop closure
measurements.
The robots poses and measurements are elements of the special Euclidean manifold
$\mathrm{SE}(d)$ where $d$ is the dimension of the problem (i.e., 2 or 3) \cite{dellaertFactorGraphsExploiting2021}.

First, assuming that the measurements noises are uncorrelated, we can
factorize eq.~\ref{eq:cslam-problem} as follows:

\begin{equation}
    \begin{aligned}
    (X_{\alpha}^{*}, X_{\beta}^{*}) \doteq 
     \argmax_{{X_{\alpha},X_{\beta}}} p(&{Z_{\alpha},Z_{\beta},Z_{\alpha\beta}}|{X_{\alpha},X_{\beta}}) \\
    \doteq 
     \argmax_{{X_{\alpha},X_{\beta}}} \bigg(
    &\prod_{i=1}^{l}p(z_{\alpha}^i|X_{\alpha}^{i})
    \prod_{j=1}^{m}p(z_{\beta}^j|X_{\beta}^{j})  
    \\ & \prod_{k=1}^{n}p(z_{\alpha\beta}^k|X_{\alpha}^{k},X_{\beta}^{k}) \bigg)
    \end{aligned}
\end{equation}
where $p(z_{\alpha}^{i}|X_{\alpha}^{i})$ is the likelihood of the $i^{th}$
measurement of robot $\alpha$ (i.e., $z_{\alpha}^{i}$) given the subset of variables
$X_{\alpha}^{i}$ on which it is dependent,
$p(z_{\beta}^{j}|X_{\beta}^{j})$ is the likelihood of the $j^{th}$
measurement of robot $\beta$ (i.e., 
$z_{\beta}^{j}$) given the subset of variables $X_{\beta}^{j}$ on which
it is dependent, and
$p(z_{\alpha \beta}^{k}|X_{\alpha}^{k},X_{\beta}^{k})$ is the
likelihood of the $k^{th}$ inter-robot measurement (i.e., $z_{\alpha \beta}^{k}$) given the subset of
variables $X_{\alpha}^{k}$ and $X_{\beta}^{k}$. There are $l$
measurements related only to state variables from robot $\alpha$, $m$
measurements related only to state variables from robot $\beta$, and
$n$ measurements related to state variables from both robots.

Second, assuming that the measurements are disturbed by zero-mean
Gaussian noise with information matrix $\Omega$ (i.e., inverse of the
covariance), we can express the individual measurement likelihood as

\begin{equation}
p(z_{\alpha}^i|X_{\alpha}^{i}) \propto \text{exp}\left(-\dfrac{1}{2}\norm{h_{\alpha}^{i}(X_{\alpha}^{i})-z_{\alpha}^{i}}_{\Omega_{\alpha}^{i}}^{2}\right)
\end{equation}
where $h_{\alpha}^{i}$ is a function that maps the state variables to
the measurements.

Third, since maximizing the likelihood is equivalent to minimizing the
negative log-likelihood, we obtain the following nonlinear least
squares formulation of problem \ref{eq:cslam-problem}:

\begin{equation}
    \begin{aligned}
        (X_{\alpha}^{*}, X_{\beta}^{*}) \doteq 
         \argmin_{{X_{\alpha},X_{\beta}}} -\text{log}\bigg(
        &\prod_{i=1}^{l}p(z_{\alpha}^i|X_{\alpha}^{i})
        \prod_{j=1}^{m}p(z_{\beta}^j|X_{\beta}^{j})  
        \\ &  \prod_{k=1}^{n}p(z_{\alpha\beta}^k|X_{\alpha}^{k},X_{\beta}^{k}) \bigg)
        \\  \doteq \argmin_{{X_{\alpha},X_{\beta}}} \bigg( \sum_{i=1}^{l}&\norm{h_{\alpha}^{i}(X_{\alpha}^{i})-z_{\alpha}^{i}}_{\Omega_{\alpha}^{i}}^{2} +
        \sum_{j=1}^{m}\norm{h_{\beta}^{j}(X_{\beta}^{j})-z_{\beta}^{j}}_{\Omega_{\beta}^{j}}^{2}
        \\ + \sum_{k=1}^{n}&\norm{h_{\alpha\beta}^k(X_{\alpha}^{k},X_{\beta}^{k})-z_{\alpha\beta}^k}_{\Omega_{\alpha\beta}^k}^{2} \bigg)
    \end{aligned}
    \label{eq:cslam-nonlinLS}
\end{equation}

This nonlinear least squares problem can be solved either on a single
computer or in a distributed fashion. In the centralized case, one can
simply use single-robot pose graph optimization solvers
\cite{agarwalCeresSolverLarge,kummerleG2oGeneralFramework2011,f.dellaertetal.GeorgiaTechSmoothing,rosenSESyncCertifiablyCorrect2019}.
Incremental single-robot solvers~\cite{kaessISAM2IncrementalSmoothing2011} can also be adapted for
the centralized \CSLAM problem to continuously update the global pose graph with
the latest measurements from the robots
\cite{dongDistributedRealtimeCooperative2015}.
Recently, a client-server architecture has been proposed in which
resource-limited clients (e.g. robots or mobile phones) only optimize small parts of the map
while the server processes the
rest~\cite{zhangDistributedClientServerOptimization2021}. This centralized and
distributed approach allows
for accurate real-time state estimation even with limited computation and memory
capacity onboard the clients.

Among the distributed solvers, many early techniques used Gaussian
elimination
\cite{cunninghamDDFSAMFullyDistributed2010,cunninghamDDFSAMConsistentDistributed2013,cunninghamFullyDistributedScalable2012}.
Although popular, those approaches require the exchange of dense
marginals which means that the communication cost is quadratic on the
number of inter-robot measurements.  
Furthermore, those approaches
rely on linearization, so they require complex bookkeeping to ensure
the consistency at the linearization point within the team of robots.
To reduce the complexity,
\cite{nerurkarDistributedMaximumPosteriori2009} introduces a
distributed marginalization scheme to limit the size of the
optimization problem.

More recently, the approach in
\cite{choudharyDistributedMappingPrivacy2017} leverages the
Distributed Gauss-Seidel algorithm introduced in
\cite{bertsekasParallelDistributedComputation1989} to solve
eq.~\ref{eq:cslam-nonlinLS}.  This technique avoids complex
bookkeeping and information double-counting in addition of satisfying
privacy constraints by exchanging minimal information on the robot
individual trajectories.
In another line of work, \cite{zhangMRiSAM2IncrementalSmoothing2021} extends a
single-robot incremental solver~\cite{kaessISAM2IncrementalSmoothing2011} towards distributed
multi-robot setups. 
This is useful in online
missions as it can update the current estimate based on the latest observations
without recomputing the whole problem.

Optimization on Riemannian manifolds~\cite{boumal2020intromanifolds} has also been considered extensively to
solve the \CSLAM problem
\cite{knuthCollaborative3DLocalization2012,knuthCollaborativeLocalizationHeterogeneous2013}.
Approaches in
\cite{tronDistributedImagebased3D2009,tronDistributed3DLocalization2014,tronDistributedOptimizationFramework2016}
introduce a multi-stage distributed Riemannian consensus protocol with
convergence guarantees to globally optimal solutions in noiseless
scenarios.  
Expanding on those ideas, a recent technique
\cite{tianDistributedCertifiablyCorrect2021}, based upon a sparse
semidefinite relaxation, provides exactness guarantees even in the
presence of moderate measurement noise.  Moreover, this latter
technique has been extended to consider asynchronous scenarios and
parallel computation \cite{tianAsynchronousParallelDistributed2020},
which are often critical to deal with communication delays inherent to
multi-robot systems.

\subsection{Other Estimation Techniques}

Other estimation techniques have been proposed for \CSLAM. Among them,
the distributed Jacobi approach has been shown to work for 2D poses
\cite{araguesMultiagentLocalizationNoisy2011}.
\cite{franceschelliAgreementProblemsGossip2010,araguesDistributedConsensusRobot2012}
look into consensus-based algorithms and prove their convergence
across teams of robots.  Also, apart from the solver itself,
researchers have studied which measurement and noise models are the
best suited for \CSLAM
\cite{indelmanGraphbasedDistributedCooperative2012}.

We observe that more exciting new directions are still being
discovered, considering that recent approaches such as
\cite{tianDistributedCertifiablyCorrect2021} have been shown to
outperform, both in accuracy and convergence rate, the well
established Distributed Gauss-Seidel pose graph optimization method
\cite{choudharyDistributedMappingPrivacy2017} reused in many
state-of-the-art \CSLAM systems such as
\cite{cieslewskiDataEfficientDecentralizedVisual2018,lajoieDOORSLAMDistributedOnline2020,wangActiveRendezvousMultiRobot2019}.
Those promising approaches also include the majorization-minimization
technique from \cite{fanMajorizationMinimizationMethods2020a}, the
consensus-based 3D pose estimation technique inspired by distributed
formation control from
\cite{cristofaloConsensusbasedDistributed3D2019,cristofaloGeoDConsensusbasedGeodesic2020},
and
\cite{zhuDistributedVisualInertialCooperative2021} distributed estimator based
on covariance intersection.

\subsection{Perceptual Aliasing Mitigation}

As it is the case in single robot \SLAM, loop closure detection is
vulnerable to spurious measurements, i.e., outliers, due to perceptual
aliasing.
This phenomenon occurs when two different places are conflated as the
same during the place recognition process.  
This motivates the need
for robust techniques that can detect and remove those outliers to
avoid dramatic distortions in the \CSLAM estimates.  
A common approach is to adopt a robust objective function less sensitive to
outliers~\cite{sunderhaufSwitchableConstraintsRobust2012,agarwalRobustMapOptimization2013,latifRobustLoopClosing2013,lajoieModelingPerceptualAliasing2019,yangGraduatedNonConvexityRobust2020}.
Outliers
mitigation might also help against adversarial attacks by rejecting
spurious measurements injected by a nefarious agent.

The classic approach to remove outliers is to use the RANSAC algorithm
\cite{fischlerRandomSampleConsensus1981} to find a set of mutually
consistent measurements
\cite{dongDistributedRealtimeCooperative2015}. While RANSAC works well
in centralized settings, it is not adapted to distributed systems.
Therefore, researchers recently explored other ways of detecting
outliers such as leveraging extra information from the wireless
communication channels during a rendezvous between two robots
\cite{wangActiveRendezvousMultiRobot2019}.  
Since such approaches work
only for direct inter-robot loop closures, there is a need for general
robust data association in the back-end.  To that end,
\cite{indelmanMultirobotPoseGraph2014} uses expectation maximization
to infer which inter-robot measurements are inliers and which ones are
outliers. 
One currently popular approach in
\CSLAM is the use of pairwise consistency maximization to search for
the maximal clique of pairwise consistent measurements among the
inter-robot loop closures
\cite{mangelsonPairwiseConsistentMeasurement2018}.
\cite{lajoieDOORSLAMDistributedOnline2020} introduces a distributed
implementation of this technique which does not require any additional
communication when paired with distributed pose graph optimization,
while \cite{changKimeraMultiSystemDistributed2021} proposes an
incremental version, and \cite{doRobustLoopClosure2020} extends the
pairwise consistency evaluation with a data similarity metric.  
Another recent work~\cite{tianKimeraMultiRobustDistributed2021} extends to
distributed setups the Graduated Non-Convexity approach for outlier rejection
previously applied to single-robot \SLAM~\cite{yangGraduatedNonConvexityRobust2020}.
It is
important to note that those latest approaches only apply to
smoothing-based \CSLAM since, unlike filtering, it allows the removal
of past measurements from the estimation.

\section{System-Level Challenges}
\label{sec:system-challenges}

Along with the front-end and back-end specific challenges, some issues and
design choices affect the whole \CSLAM system. As in single-robot \SLAM, the map
representation has strong repercussions on motion tracking, place recognition
and state estimation. On top of this, multi-robot systems (described in
\ref{sec:multi-rob}) present unique challenges to \CSLAM in terms of
communication and coordination.

\subsection{Map Representation}
\label{sec:map}

When designing large multi-robot systems, the choices of map representation could
affect computation load, memory usage, and communication bandwidth.
First, it is important to note that an interpretable map is not always
required.
For example, when the sole objective is collaborative localization, a feature
map can be sufficient. In those cases, each robot locally tracks landmarks, or
features, and searches for correspondences in other robots' feature maps to
obtain indirect inter-robot loop closure measurements. The local feature maps
can be merged frequently so that the robots can navigate and track features in a
global map, or they can be shared on demand upon place recognition events. This
way, the robots can operate in the same reference frame without the computation
and communication burden of building an interpretable map model.

When required, the chosen map representation
depends on the mission objective and environment. For example, in the case of
ground robots in flat indoor environments, a 2D map might be
sufficient~\cite{caccavaleWireframeMap}. In those scenarios, occupancy grid maps
have been shown to be a compact and more accurate solution
\cite{martinJustintimeCooperativeSimultaneous2010,saeediMultipleRobotSimultaneous2011}
than feature-based maps \cite{benedettelliMultirobotSLAMUsing2010}. However, 3D
representations are sometimes required (e.g. for rough terrain navigation)
at the cost of more computation, storage, and communication, which can
be difficult to handle when resources are limited on the robots.
Given the communication constraints in \CSLAM systems, compact or
sparse representations, such as topological maps
\cite{h.jackychangMultirobotSLAMTopological2007,saeediGroupMappingTopological2014},
are often preferred.
In the same vein, some works aim for semantic-based representations in
the form of sparse maps of labelled regions
\cite{choudharyMultiRobotObjectBased2017}.  Map representations can
also affect long-term operations due to the increasing size of the map
in memory \cite{zhangDistributedCollaborativeMonocular2018}, which is
also a challenge in single-robot \SLAM.

\subsection{Efficient and Robust Communication}
\label{sec:communication}
One of the core implementation differences between \SLAM and \CSLAM is the need for
communication and coordination within the robotic team. 
For efficiency, the required bandwidth needs to be minimal, and the
communication network needs to be robust to robot failures and varying topologies.

The communication bottleneck of a \CSLAM system is generally caused by the
exchanges of sensor data or representations used to compute inter-robot loop
closures \cite{tardioliVisualDataAssociation2015}. Robots need to share enough
data to detect whether other robots have visited the same area, and then
estimate a map alignment using any overlapping sections of their maps. Hence,
contributions to the front-end of \CSLAM systems often consist of mechanisms to
perform an efficient search for loop closure candidates over a team, considering
communication constraints. Conversely, the back-end generally relies on a
pose-graph which can be shared without the need for large bandwidth.

\subsubsection{Efficient Data Sharing}
While some early techniques simply share all the data from one robot to another,
new sensors produce increasingly rich and dense data. The days of raw sensor
data transmission are over and most current techniques in literature opt for
some sort of compression or reduction. Even among the early techniques
\cite{nettletonDecentralisedSLAMLowBandwidth2006}, the idea of a communication
budget has been explored. More recently, the topic has gathered more attention
with new techniques carefully coordinating the exchange of data when two robots
meet, accounting for the available communication and computation resources
\cite{giamouTalkResourceEfficientlyMe2018,tianNearOptimalBudgetedData2018,tianResourceAwareAlgorithmsDistributed2018,tianResourceawareApproachCollaborative2020}.
One idea is to compress the generated maps using self-organizing maps obtained
through unsupervised learning
\cite{saeediNeuralNetworkBasedMultiple2011,bestDecentralisedSelfOrganisingMaps2020}.
The use of compact representations has also been explored with high-level
semantic features: \cite{choudharyMultiRobotObjectBased2017} relies on objects
as landmarks, needing to communicate only object labels and poses to other
robots, and \cite{ramtoulaCAPRICORNCommunicationAware2020} presents a compact
object-based descriptor relying on the configuration of objects in a scene to
perform place recognition. In addition to making representations compact, it is
useful to ensure that only helpful information is shared. Hence,
\cite{keplerApproachReduceCommunication2020} introduces a novelty metric so that
only sufficiently novel measurements compared to the existing map are
transmitted.

The problem has been extensively studied specifically for visual \CSLAM:
\cite{tardioliVisualDataAssociation2015} proposes to share visual
vocabulary indexes instead of feature descriptors to reduce the
required bandwidth. 
Other approaches focus on scalable team-wide place recognition by assigning each
robot with a predetermined range of words from a pretrained visual bag of
words \cite{cieslewskiEfficientDecentralizedVisual2017}, or regions of
full-image descriptors \cite{cieslewskiEfficientDecentralizedVisual2017a}.
\cite{dymczykGistMapsSummarizing2015,contrerasOPOCOOnlinePoint2017}
remove landmarks that are not necessary for localization,
\cite{opdenboschCollaborativeVisualSLAM2019} introduces a new coding
to efficiently compress features, and
\cite{duboisDataSharingStrategy2019} proposes data sharing algorithms
specialized for visual inertial \CSLAM.

In some extreme cases, communication is severely limited due to the properties
of the transmission medium or the large distance between the robots:
\cite{paullDecentralizedCooperativeTrajectory2014,paullCommunicationconstrainedMultiAUVCooperative2015}
explore the special case of underwater operations with low bandwidth acoustic
communication, and \cite{schulzSimultaneousCollaborativeMapping2019} considers
long distance radio modules with very limited bandwidth to build the
collaborative map through small incremental updates.

\subsubsection{Network Topology}
Another important aspect to consider is the network topology. Current
techniques either assume full connectivity, multi-hop connectivity or
are rendezvous-based.  Full connectivity means that each robot can
directly communicate with all other robots at any time such as in
\cite{cieslewskiEfficientDecentralizedVisual2017a,cieslewskiEfficientDecentralizedVisual2017}.
Multi-hop connectivity implies that robots can only share information
with their neighbors and it might require multiple
neighbor-to-neighbor transmissions to reach all robots
\cite{araguesConsistentDataAssociation2010,montijanoDistributedDataAssociation2013}.
Rendezvous-based communication means that the robots share data only
when they meet and, therefore, do not require any connectivity
maintenance.  Rendezvous-based \CSLAM also offers the opportunity to
perform direct inter-robot loop closure detection during the
encounters \cite{zhouMultirobotSLAMUnknown2006}.

The impact of the network topology is especially important during the
inference step because disconnections or multi-hop paths can lead to
inconsistencies or synchronization issues. Thus,
\cite{leungDistributedDecentralizedCooperative2011,leungDecentralizedCooperativeSLAM2012}
examine the conditions that allow distributed inference to reach the
same result as a centralized equivalent approach. 
Another approach
\cite{quraishiRobustnessConnectivityLoss2016} leverages the progress
in the field of distributed computing to improve the robustness to
connectivity losses, while \cite{tunaWirelessSensorNetworkbased2015a}
evaluates the use of \textit{Wireless Sensor Network}-based
communication which is less reliable and predictable, but offers a
flexible architecture with self-organization capabilities.

\section{Benchmarking \CSLAM}
\label{sec:benchmarking}
Despite the tremendous progress in the field during the last decade,
\CSLAM techniques face tough challenges in terms of reproducibility
and benchmarking.  \CSLAM systems involve multiple software modules
and lots of different hardware components, making it hard to replicate
perfectly.  While standardized benchmarking approaches have been
emerging for single-robot \SLAM
\cite{bujancaSLAMBenchSystematicAutomated2019}, such systematic
evaluation techniques are still lacking for \CSLAM.

Moreover, only a few datasets dedicated to \CSLAM exist.
\cite{leungUTIASMultirobotCooperative2011} consists of 9 monocular camera
subdatasets and \cite{duboisAirMuseumHeterogeneousMultirobot2020} is dedicated
to stereo-inertial \CSLAM. Therefore, the common approach to evaluate \CSLAM
solutions is to split single robot \SLAM datasets into multiple parts and to
associate each one to a robot. When splitting the dataset, careful attention has
to be given to ensure the presence of overlaps between the parts for loop
closing. In addition, one should avoid overlaps near the cutting points, where
the viewpoint and lighting conditions are exactly the same since they depict the
same place viewed by the robot at the same point in time: this kind of overlaps
is highly unrealistic in multi-robot operations. One of the most used dataset in
the literature is the KITTI self-driving car dataset comprised of lidar and
stereo camera data \cite{geigerAreWeReady2012}. New datasets of interest include
KITTI360 \cite{xieSemanticInstanceAnnotation2016} which adds fish-eye cameras,
the very large Pit30M lidar and monocular camera dataset that contains over 30
million frames \cite{martinezPit30MBenchmarkGlobal2020}, and the DARPA SubT
datasets which come with standardized evaluation tools for
SLAM~\cite{rogersTestYourSLAM2020,rogersSubtUrban}.

\section{Open Problems and Ongoing Trends}
\label{sec:trends}
This section presents open problems and trending ideas in the research community to
improve \CSLAM. These new trends push the boundaries of what \CSLAM
can do and offer an exciting view of the field's future.

\subsection{Resilient Inter-Robot Communication}
\label{future-comm}

Although the limitations of inter-robot communication have been a major concern
since the inception of \CSLAM, it is still one of the main open problems in the
field. In particular, there is a need for resilient communication strategies,
aiming beyond robustness to endure unexpected disruptions and ensure swift
recovery~\cite{prorokRobustnessTaxonomyApproaches2021}. Delays and dropouts are
inevitable in realistic systems, and their effects are amplified when multiple
robots operating simultaneously are flooding the network. Delays and
out-of-sequence messages can have dramatic effects on real-time robot control
which heavily relies on accurate and up-to-date state estimates from
\CSLAM~\cite{bressonSimultaneousLocalizationMapping2017}, and yet they still
have not been thoroughly addressed by the research community. Instead, current
approaches focus primarily on minimizing communication, which can be achieved,
for example, by posing distributed loop closure detection as an optimization
problem subject to a budget constraint on total data
transmission~\cite{tianNearOptimalBudgetedData2018}.

Another open problem inherent to \CSLAM and inter-robot communication is the
risk of adversarial attacks. In a future in which robots, such as autonomous
cars, collaborate on a large scale, security and data integrity will be one of
the major concerns of consumers. In addition to the usual risks of infection and
hijacking, byzantine data manipulation could lead to map merging poisoning and
intentionally 
erroneous \CSLAM estimates~\cite{dengInvestigationByzantineThreats2021}. 
Thus, further investigation and more efforts have to be deployed on system security.

An interesting, yet still underdeveloped, trend is to leverage the communication
medium for inter-robot measurements. This has been successfully done with
UWB~\cite{gentnerCooperativeSimultaneousLocalization2018,borosonInterRobotRangeMeasurements2020,caoVIRSLAMVisualInertial2021}
or WiFi~\cite{liuCollaborativeSLAMBased2020}, and could be a promising avenue
using multipath analysis with channel estimators in 5G
networks~\cite{ge5GSLAMLowcomplexity2021}. Future techniques might even sidestep
inter-robot data transfer completely by communicating via sensor observations
of each other and predetermined cues such as visual tags or behavioral
patterns~\cite{kimMultipleRelativePose2010}.

\subsection{Managing Limited Computing Resources}

Aside from communication, computational constraints are an essential
consideration in robotics since robots are usually equipped with limited onboard
processing devices. It is particularly important in \CSLAM where multiple
sub-processes from sensory analysis to inter-robot communication need to be run
simultaneously. Thus, to support the current expansion of \CSLAM capabilities,
there is a constant need for efficiency gains. In fact, computation improvements
are often at the forefront of new trends in \CSLAM such as the rise of semantic
methods, discussed in Section~\ref{sec:semantics}, which were enabled by
GPU-based deep learning~\cite{krizhevskyImageNetClassificationDeep2012}.
Moreover, as discussed in Section~\ref{sec:augmented}, many new applications of
\CSLAM are designed for even smaller platforms such as mobile phones.

Centralized techniques are a natural solution to limited onboard computation
capabilities, and, in that regard, recent research suggest that \CSLAM could
efficiently leverage the progress in cloud computing. The connection between the
two fields is somewhat intuitive: why perform all the processing on robots with
limited resources when we could use powerful remote clusters of servers instead?
For example, \cite{riazueloC2TAMCloudFramework2014} offloads the expensive map
optimization and storage to a server in the cloud.
\cite{yunCloudRoboticsPlatform2017} proposes a cloud robotics framework for
\CSLAM based on available commercial platforms. Using a similar approach,
\cite{zhangCloudBasedFrameworkScalable2018} manage to perform \CSLAM with up to
256 robots. This is orders of magnitude more than the current techniques based
on onboard computation can achieve.

However, while cloud techniques solve the problem of limited computing power
onboard the robots, they still face the issue of limited communication bandwidth
which is exacerbated when many robots transmit their data through a single
communication link. Hence, instead of using remote servers, other strategies need to
be explored. For example, a subset of a team of robots could act as a computing
cluster to free other robots from the heavy computation
burden~\cite{gouveiaComputationSharingDistributed2015}. Such moving clusters
performing computing closer to the sources of data are in accordance with the
edge computing paradigm \cite{satyanarayananEmergenceEdgeComputing2017} to save
bandwidth and reduce response time~\cite{huangEdgeRoboticsEdgeComputingAccelerated2021}.

\subsection{Adapting to Dynamic Environments}

Another inherent problem in multi-robot system is the presence of
moving objects in the environment (e.g. people or vehicles). 
In this regard, the other moving robots in the team are especially problematic.
This is a substantial issue since \SLAM techniques
rely on the tracking of static landmarks.
Attempting to solve this problem, \cite{leeMultirobotSLAMUsing2009} proposes the
simple idea of pointing the cameras towards the ceiling when operating indoors
with ground robots so that they cannot see each other.
\cite{zouCoSLAMCollaborativeVisual2013} proposes instead to classify dynamic
points using the reprojection error and to keep only the static points for
estimation. In a different vein,
\cite{moratuwageCollaborativeMultivehicleSLAM2013,moratuwageRFSCollaborativeMultivehicle2014,battistelliRandomSetApproach2017}
and more recently \cite{gaoRandomFiniteSetBasedDistributedMultirobot2020} extend
upon the Rao-Blackwellized particle filters framework to track moving features,
potentially neighboring robots, and remove them from the estimation process.
Those works use \textit{Random-Finite-Sets} which were originally developed for
multi-target tracking. This way, they manage to incorporate data association,
landmark appearance and disappearance, missed detections, and false alarms in
the filtering process. Nevertheless, handling dynamic landmarks remains an open
topic given that most current works still rely on static environment
assumptions.

\subsection{Active \CSLAM}\label{sec:active}

The concept of active \CSLAM comes from the powerful idea that while \CSLAM
naturally improves path planning and control, path planning and control can also
improve \CSLAM. Interestingly, some of the earliest works in collaborative
localization were already leveraging coordination between robots to improve
accuracy. Instead of mapping the environment, they relied on subsets of robots,
in alternance, to serve as landmarks for the
others~\cite{kurazumeCooperativePositioningMultiple1994,trawnyOptimizedMotionStrategies2004}.
In an interesting turn of events, the recent progress in \CSLAM has brought back
this active sensing trend to the forefront of research.

In \CSLAM, gains can be made by leveraging the coordination between the mapping
robots. Having feedback loops to the \CSLAM algorithm allows path planning
optimization for faster coverage and mapping of the environment
\cite{brysonCooperativeLocalisationMapping2007,brysonArchitecturesCooperativeAirborne2009}.
To achieve those goals, \cite{mahdouiCommunicatingMultiUAVSystem2020} aims to
minimize the global exploration time and the average travelled distance among
the robots. Other examples of the coupling between path planning and \SLAM
include \cite{trujilloCooperativeMonocularBasedSLAM2018}, which shows the
advantages of UAVs flying in formation for monocular \CSLAM, and
\cite{peiActiveCollaborationRelative2020} which uses deep Q-learning to decide
whether a robot should localize the others or continue exploring on its own.

Active \CSLAM can also increase the estimation accuracy. To that end,
\cite{dinnissenMapMergingMultiRobot2012} uses reinforcement learning to
determine the best moment to merge the local maps, and
\cite{kontitsisMultirobotActiveSLAM2013} leverages instead the covariance matrix
computed by the EKF-based inference engine to select trajectories that reduce
the map uncertainty. Similarly,
\cite{atanasovDecentralizedActiveInformation2015} develop a theoretical
approach to design a sensor control policy which minimizes the entropy of the
estimation task, while \cite{chenBroadcastYourWeaknesses2020} proposes to
broadcast the weakest nodes in the \CSLAM pose graph topology to actively
increase the estimation accuracy.

Those works are most likely the mere beginning of active \CSLAM research given
that \CSLAM systems are now being integrated on actual industrial, scientific or
consumer robots, opening many possibilities of interaction between \CSLAM and
other robotics subsystems.

\subsection{Semantic \CSLAM}
\label{sec:semantics}
With the rise of deep learning and its impressive semantic inference
capabilities, a lot of interest have been directed towards semantic
mapping in which the environment is interpreted using class labels
(i.e., person, car, chair, etc.).
Representing maps as a collection of objects or semantic classes usually leads to much
more compressed representations of the environment~\cite{salas-morenoSLAMSimultaneousLocalisation2013}, and this
can be especially beneficial for \CSLAM.
Indeed, fewer landmarks and smaller maps are better suited to tight
communication constraints since they reduce the amount of data sharing
between the robots.  

Semantic segmentation was first applied to \CSLAM in
\cite{wuCooperativeMultiRobotMonocularSLAM2009} which detects blobs of colors as
salient landmarks in the robots maps. \cite{choudharyMultiRobotObjectBased2017}
later leverages deep learning-based object detection to perform object-based
\CSLAM. However, such object-based techniques rely heavily upon the presence of
many objects of the known classes in the environment (i.e., classes in the
training data). Thus, they do not generalize well to arbitrary settings.

The other current preferred approach for semantic \CSLAM is to annotate maps of
the environment with class labels. For example,
\cite{freyEfficientConstellationBasedMapMerging2019,ramtoulaCAPRICORNCommunicationAware2020}
use constellations of landmarks each comprised of a 3D point cloud, a class
label and an appearance descriptor. The relatively small number of semantic
landmarks reduces the required inter-robot communication significantly.
\cite{tchuievDistributedConsistentMultiRobot2020a} considers the joint
estimation of object labels and poses in addition to the robots poses in order
to improve both estimates. \cite{changKimeraMultiSystemDistributed2021} build
instead globally consistent local metric maps that are enhanced with local
semantic labelling, hence preserving the accuracy of 
pure geometric \CSLAM approaches while incorporating useful high-level
information in the robots 
individual maps.

The tremendous progress still occurring in the field of deep learning strongly
suggests that there is more to come in terms of integration with \CSLAM and
enhanced collaborative understanding of the environment.
\subsection{Augmented Reality}\label{sec:augmented}

Apart from the well known UAV or self-driving cars applications, Augmented
Reality (AR) is probably one of the biggest field of application of \SLAM.
Indeed, \SLAM makes markerless AR applications possible by building a map of the
surrounding environment which is essential to overlay digital interactive
augmentations. In other words, \SLAM is required to make AR work in environments
without motion capture, localization beacons or predetermined markers. In the
foreseeable future, AR applications and games will push for multi-agent
collaboration and this is where \CSLAM comes into play
\cite{egodagamageCollaborativeAugmentedReality2017,egodagamageDistributedMonocularVisual2018}.
To that end, \cite{morrisonMOARSLAMMultipleOperator2016} proposes a centralized
approach in which virtual elements are shared by all agents, and
\cite{sartipiDecentralizedVisualInertialLocalization2019} introduces a
decentralized AR technique with smartphones, making use of the visual and
inertial sensors already present in those devices. In a similar vein,
\cite{guoResourceAwareLargeScaleCooperative2018} presents a resource-aware
technique capable of trading off accuracy to adjust the computational cost to
the available resources on mobile devices.

Some other techniques also look at the tremendous potential of collaborative AR
for intuitive human-robot interfaces which is especially complex when the number
of agents (i.e., humans or robots) and viewpoints increases. For example, to
improve supervised mapping tasks, \cite{sidaouiCollaborativeHumanAugmented2019}
equips a human operator with an AR system to edit and correct the map produced
by a robot during a mission. Interestingly,
\cite{yuCollaborativeSLAMARguided2020} goes in the opposite direction: humans
equipped with smartphones map an environment and get feedback from a central
server to indicate which unscanned areas still need to be explored.

Augmented Reality might soon become the main application of \CSLAM in our daily
lives, but there is still a lot of research work ahead to efficiently satisfy
its inherent constraints and achieve robust large-scale deployments.

\section{Conclusions}
\label{sec:conclusion}

In this paper, we presented the core ideas behind Collaborative Simultaneous
Localization and Mapping and provided a survey of existing techniques. First, we
introduced the basic concepts of a \CSLAM system. We provided explanations and
bits of historical context to better understand the astonishing progress
recently made in the field. Then, we presented the building blocks of a typical
\CSLAM system and the associated techniques in the literature. We also touched
upon the difficulties of reproducibility and benchmarking. Afterwards, we
explored new trends and challenges in the field that will certainly receive a
lot more interests in the future. In summary, we focused on providing a complete
overview of the \CSLAM research landscape.

We have shown, through numerous examples, how \CSLAM systems are varied and need
to match closely the application requirements: sparse or dense maps, precise or
topological localization, the number of robots involved, the networking
limitations, etc. We wish for this survey to be a useful tool for \CSLAM
practitioners looking for adequate solutions to their specific problems.

Nevertheless, despite the current growing interest for \CSLAM applications, it
is still a young topic of research and many fundamental problems have to be
resolved before the advance of \CSLAM-based commercial products. In particular,
we believe that current systems scale poorly and are often limited to very few
robots. So, a lot of work is still required to achieve large teams of robots
building maps and localizing themselves collaboratively. We also note the
growing interest for semantic \CSLAM to make robotic maps more interpretable and
more actionable. Scene understanding techniques in the computer vision field
could bring more compact and expressive environment representations into the
\SLAM system, which potentially increase the map readability while reducing the
inter-robot communication burden.
Furthermore, the rise of AR, in conjunction with \CSLAM and semantics, will
offer incredible opportunities of innovation in the fields of collaborative
robotics, mobile sensing, and entertainment.

\subsubsection*{Acknowledgments}
This work was partially supported by a Canadian Space Agency FAST
Grant, a Vanier Canada Graduate Scholarships Award, the Arbour
Foundation, the EPSRC Centre for Doctoral Training in Autonomous
Intelligent Machines and Systems [EP/S024050/1], and Oxbotica.

\bibliographystyle{apalike} 

\end{document}